%% file: main_arxiv.tex
\documentclass[10pt,twocolumn,letterpaper]{article}
\usepackage{cvpr}
\usepackage{times}
\usepackage{epsfig}
\usepackage{graphicx}
\usepackage{amsmath}
\usepackage{amssymb}
\usepackage{rotating}
\usepackage{algorithm}
\usepackage{algpseudocode}
\usepackage{subcaption}
\usepackage{array}
\usepackage{siunitx,booktabs}

\usepackage[pagebackref=true,breaklinks=true,letterpaper=true,colorlinks,bookmarks=false]{hyperref}

\cvprfinalcopy 

\ifcvprfinal\fi
\begin{document}

\title{Learning from Synthetic Data: Addressing Domain Shift for Semantic Segmentation}

\author{Swami Sankaranarayanan$^{1}$ \thanks{First two authors contributed equally}
\quad Yogesh Balaji $^{1}$$^{*}$ \quad Arpit Jain $^{2}$ \quad Ser Nam Lim $^{2,3}$ \quad Rama Chellappa $^{1}$ \\
\\
$^{1}$ UMIACS, University of Maryland, College Park, MD
\\
$^{2}$ GE Global Research, Niskayuna, NY 
\\
$^{3}$ Avitas Systems, GE Venture, Boston MA. 
}
\maketitle

\begin{abstract}
  Visual Domain Adaptation is a problem of immense importance in computer vision. Previous approaches showcase the inability of even deep neural networks to learn informative representations across domain shift. This problem is more severe for tasks where acquiring hand labeled data is extremely hard and tedious. In this work, we focus on adapting the representations learned by segmentation networks across synthetic and real domains. Contrary to previous approaches that use a simple adversarial objective or superpixel information to aid the process, we propose an approach based on Generative Adversarial Networks (GANs) that brings the embeddings closer in the learned feature space. To showcase the generality and scalability of our approach, we show that we can achieve state of the art results on two challenging scenarios of synthetic to real domain adaptation. Additional exploratory experiments show that our approach: (1) generalizes to unseen domains and (2) results in improved alignment of source and target distributions. 
\end{abstract}

\input{intro}
\input{background}
\input{method}

\input{results}

\input{ablation_studies}

\input{conclusion}
\input{ackno}
{
\bibliographystyle{ieee}
\bibliography{egbib}
}
\end{document}

%% file: intro.tex
\section{Introduction}
\label{sec:intro}

Deep Convolutional Neural Networks (DCNNs) have revolutionalized the field of computer vision, achieving the best performance in a multitude of tasks such as image classification~\cite{resnet2016}, semantic segmentation~\cite{fcn2015}, visual question answering~\cite{VQA_coattention} etc. This strong performance can be attributed to the availability of abundant labeled training data. While annotating data is relatively easier for certain tasks like image classification, they can be extremely laborious and time-consuming for others. Semantic segmentation is one such task that requires great human effort as it involves obtaining dense pixel-level labels. The annotation time for obtaining pixel-wise labels for a single image from the CITYSCAPES dataset is about 1 hr., highlighting the level of difficulty (\cite{Cityscapes}, \cite{GTA5}). The other challenge lies in collecting the data: While natural images are easier to obtain, there are certain domains like medical imaging where collecting data and finding experts to precisely label them can also be very expensive.

\begin{figure}
\includegraphics[width=0.5\textwidth]{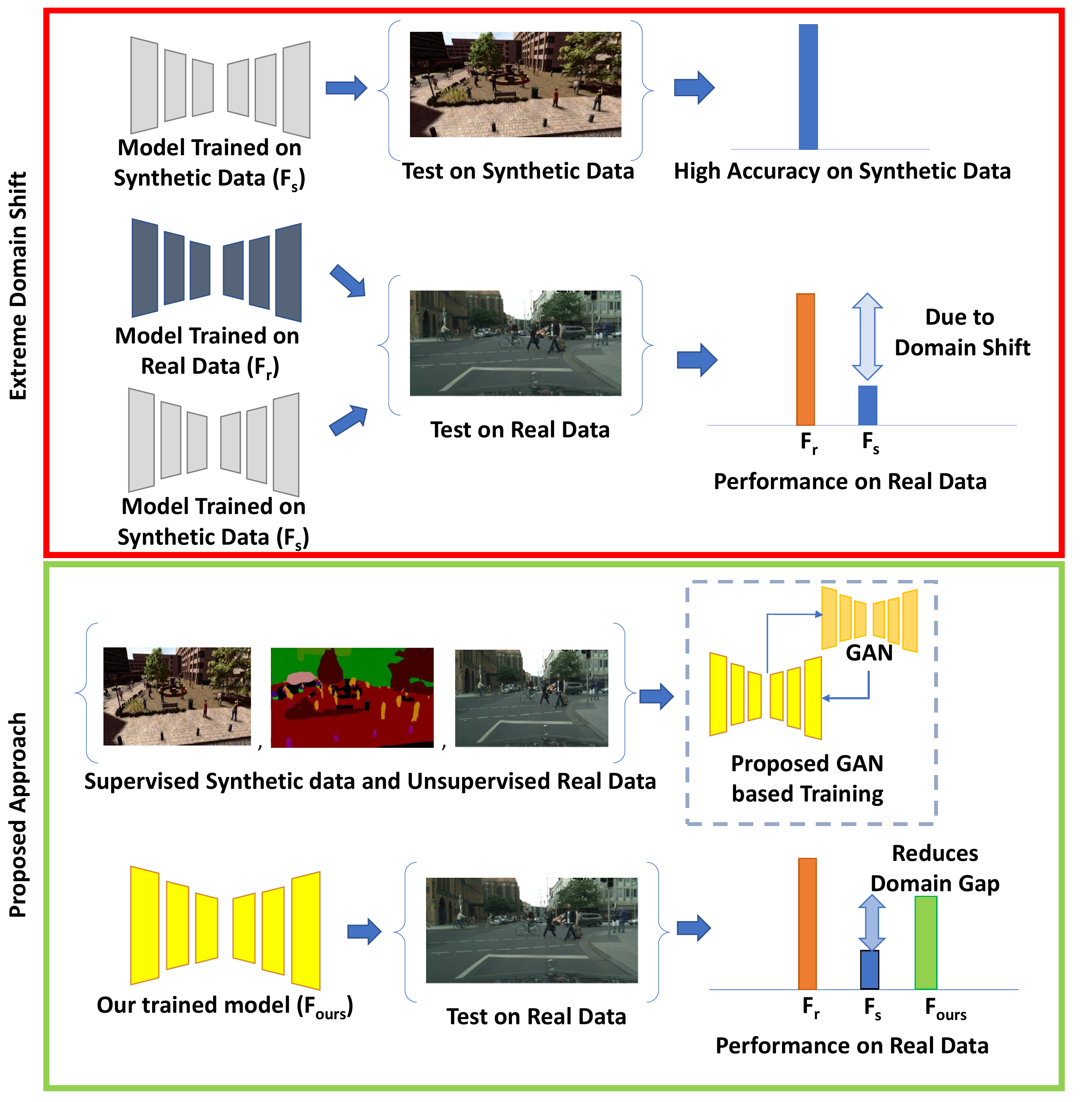}
\caption{Characterization of Domain Shift and effect of the proposed approach in reducing the same}
\label{fig:intro}
\end{figure}

One promising approach that addresses the above issues is the utility of synthetically generated data for training. However, models trained on the synthetic data fail to perform well on real datasets owing to the presence of domain gap between the datasets. Domain adaptation encompasses the class of techniques that address this domain shift problem. Hence, the focus of this paper is in developing domain adaptation algorithms for semantic segmentation. Specifically, we focus on the hard case of the problem where no labels from the target domain are available. This class of techniques is commonly referred to as Unsupervised Domain Adaptation.

Traditional approaches for domain adaptation involve minimizing some measure of distance between the source and the target distributions. Two commonly used measures are Maximum Mean Discrepancy (MMD) (~\cite{mmd2011}, ~\cite{DAN}~\cite{RTN}), and learning the distance metric using DCNNs as done in Adversarial approaches (~\cite{ganin2014},~\cite{ADDA}). Both approaches have had good success in the classification problems; however, as pointed out in \cite{curriculum_DA}, their performance improvement does not translate well to the semantic segmentation problem. This motivates the need for developing new domain adaptation techniques tailored to semantic segmentation.

The method we present in this work falls in the category of aligning domains using an adversarial framework. Among the recent techniques that address this problem,  FCN in the wild \cite{FCN_wild} is the only approach that uses an adversarial framework.  However, unlike \cite{FCN_wild} where a discriminator operates directly on the feature space, we project the features to the image space using a generator and the discriminator operates on this projected image space. Adversarial losses are then derived from the discriminator. We observed that applying adversarial losses in this projected image space achieved a significant performance improvement as compared to applying such losses directly in the feature space (ref. Table \ref{tab:ablation}). 

The main contribution of this work is that we propose a technique that employs generative models to align the source and target distributions in the feature space. We first project the intermediate feature representations obtained using a DCNN to the image space by training a reconstruction module using a combination of $L_{1}$ and adversarial losses. We then impose the domain alignment constraint by forcing the network to learn features such that source features produce target-like images when passed to the reconstruction module and vice versa. This is accomplished by employing a series of adversarial losses. As training progresses, the generation quality gradually improves, while at the same time, the features become more domain invariant.



%% file: background.tex
\section{Related Work}\label{sec:background}
Fully Convolutional Networks (FCN) by Shelhamer \emph{et al} \cite{fcn2015} signified a paradigm shift in how to fully exploit the representational power of CNNs for the semantic pixel labeling tasks. While performance has been steadily improving for popular benchmarks such as PASCAL VOC~\cite{PASCAL} and MS-COCO~\cite{MSCOCO}, they do not address the challenges of domain shift within the context of semantic segmentation.

Domain adaptation has been widely explored in computer vision primarily for the classification task. Some of the earlier approaches involved using feature reweighting techniques~\cite{Daume_DA}, or constructing intermediate representations using manifolds (~\cite{Gopalan_DA},~\cite{GFK}). Since the advent of deep neural networks, emphasis has been shifted to learning domain invariant features in an end-to-end fashion. A standard framework for deep domain adaptation involves minimizing a measure of domain discrepancy along with the task being solved. Some approaches use Maximum Mean Discrepancy and its kernel variants for this task (~\cite{DAN}, ~\cite{RTN}), while others use adversarial approaches (~\cite{ganin2014},~\cite{PixelDA},~\cite{sankaranarayanan2017generate}).

We focus on adversarial approaches since they are more related to our work. Revgrad~\cite{ganin2014} performs domain adaptation by applying adversarial losses in the feature space, while PixelDA~\cite{PixelDA} and CoGAN~\cite{CoGAN} operate in the pixel space. While these techniques perform adaptation for the classification task, there are very few approaches aimed at semantic segmentation. To the best of our knowledge, ~\cite{FCN_wild} and ~\cite{curriculum_DA} are the only two approaches that address this problem. FCN in the wild ~\cite{FCN_wild} proposes two alignment strategies - (1) global alignment which is an extension to the domain adversarial training proposed by ~\cite{ganin2014} to the segmentation problem and (2) local alignment which aligns class specific statistics by formulating it as a multiple instance learning problem. Curriculum domain adaptation~\cite{curriculum_DA} on the other hand proposes curriculum-style learning approach where the easy task of estimating global label distributions over images and local distributions over landmark super-pixels is learnt first. The segmentation network is then trained so that the target label distribution follow these inferred label properties.


One possible direction to address the domain adaptation problem is to employ style transfer or cross domain mapping networks to stylize the source domain images as target and train the segmentation models in this stylized space. Hence, we discuss some recent work related to the style transfer and unpaired image translation tasks. The popular work of Gatys \emph{et al.}~\cite{gatys_style_transfer} introduced an optimization scheme involving backpropagation for performing content preserving style transfer, while Johnson \emph{et al.}~\cite{johnshon_style_transfer} proposed a feed-forward method for the same. CycleGAN~\cite{cycleGAN_ICCV} performs unpaired image-to-image translation by employing adversarial losses and cycle consistency losses. In our experiments, we compare our approach to some of these style-transfer based data augmentation schemes.


%% file: method.tex
\section{Method}\label{sec:method}

In this section, we provide a formal treatment of the proposed approach and explain in detail our iterative optimization procedure. Let $X \in \mathbb{R}^{M \times N \times C}$ be an arbitrary input image (with $C$  channels) and $Y \in \mathbb{R}^{M \times N}$ be the corresponding label map.  Given an input $X$, we denote the output of a CNN as $\hat{Y} \in \mathbb{R}^{M \times N \times {N_c}}$, where $N_c$ is the number of classes.  $\hat{Y}(i,j) \in \mathbb{R}^{N_c}$ is a vector representing the class probability distribution at pixel location $(i,j)$ output by the CNN. The source(s) or target (t) domains are denoted by a superscript such as $X^s$ or $X^t$. 

First, we provide an input-output description of the different network blocks in our pipeline. Next, we describe separately the treatment of source and target data, followed by a description of the different loss functions and the corresponding update steps. Finally, we motivate the design choices involved in the discriminator ($D$) architecture.

\subsection{Description of network blocks}
Our training procedure involves alternatively optimizing the following network blocks:

(a) The base network, whose architecture is similar to a pre-trained model such as VGG-16, is split into two parts: the embedding denoted by $F$ and the pixel-wise classifier denoted by $C$. The output of $C$ is a label map up-sampled to the same size as the input of $F$. 

(b) The generator network ($G$) takes as input the learned embedding and reconstructs the RGB image.

(c) The discriminator network ($D$) performs two different tasks given an input: (a) It classifies the input as real or fake in a domain consistent manner (b) It performs a pixel-wise labeling task similar to the $C$ network. Note that (b) is active only for source data since target data does not have any labels during training.

\begin{figure}
\includegraphics[width=0.45\textwidth]{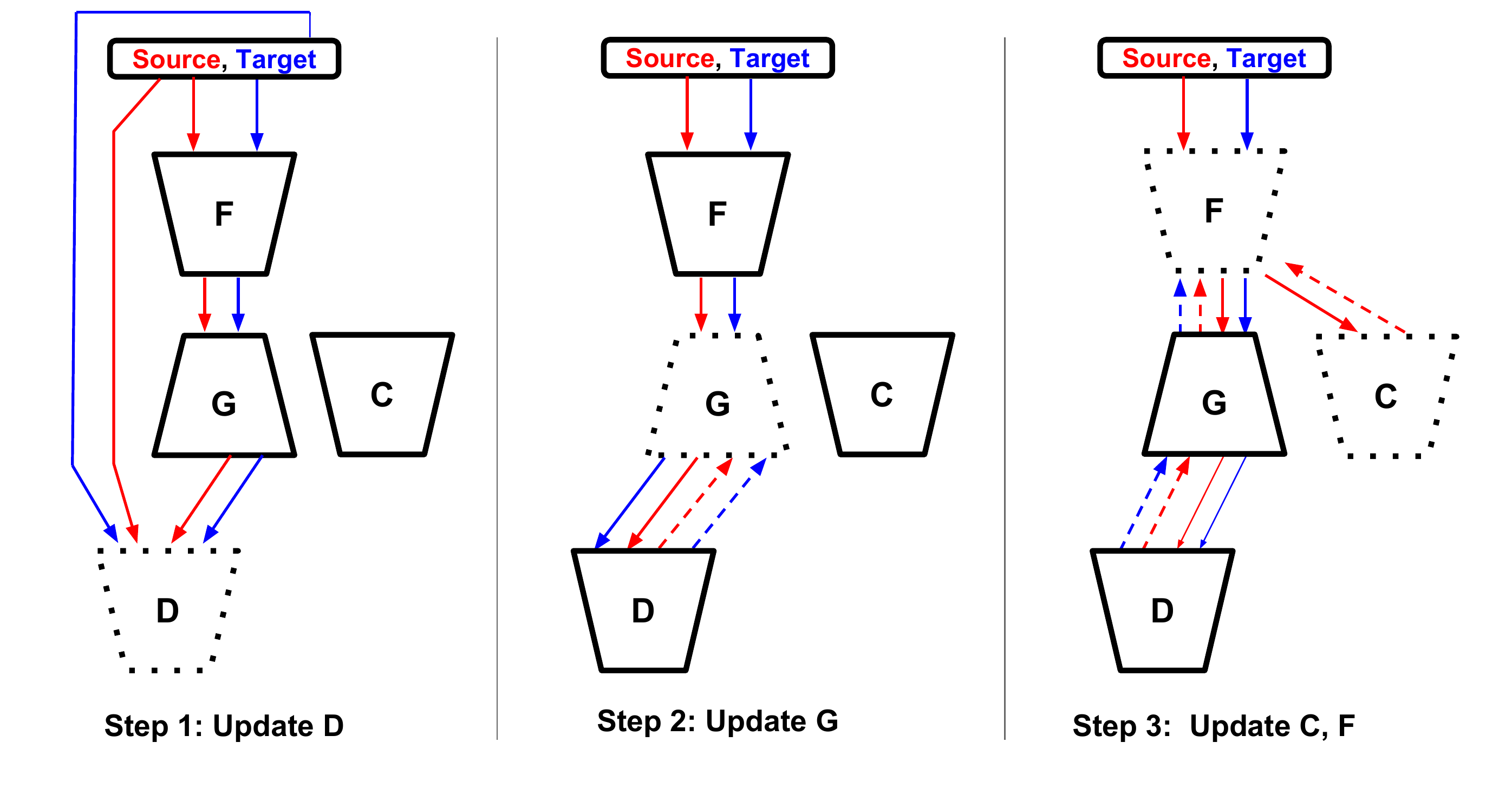}
\caption{The directions of data flow \textit{solid arrows} during the forward pass and gradient flow \textit{dotted arrows} during the backward pass of our iterative update procedure. \textit{Solid} blocks indicate that the block is frozen during that update step while \textit{dotted} block indicate that it is being updated. {\color{red} Red} denoted source information and {\color{blue} Blue} denotes target information.}
\label{fig:explain}
\end{figure}

\subsection{Treatment of source and target data}
\begin{figure*}
\centering
\includegraphics[scale=0.28]{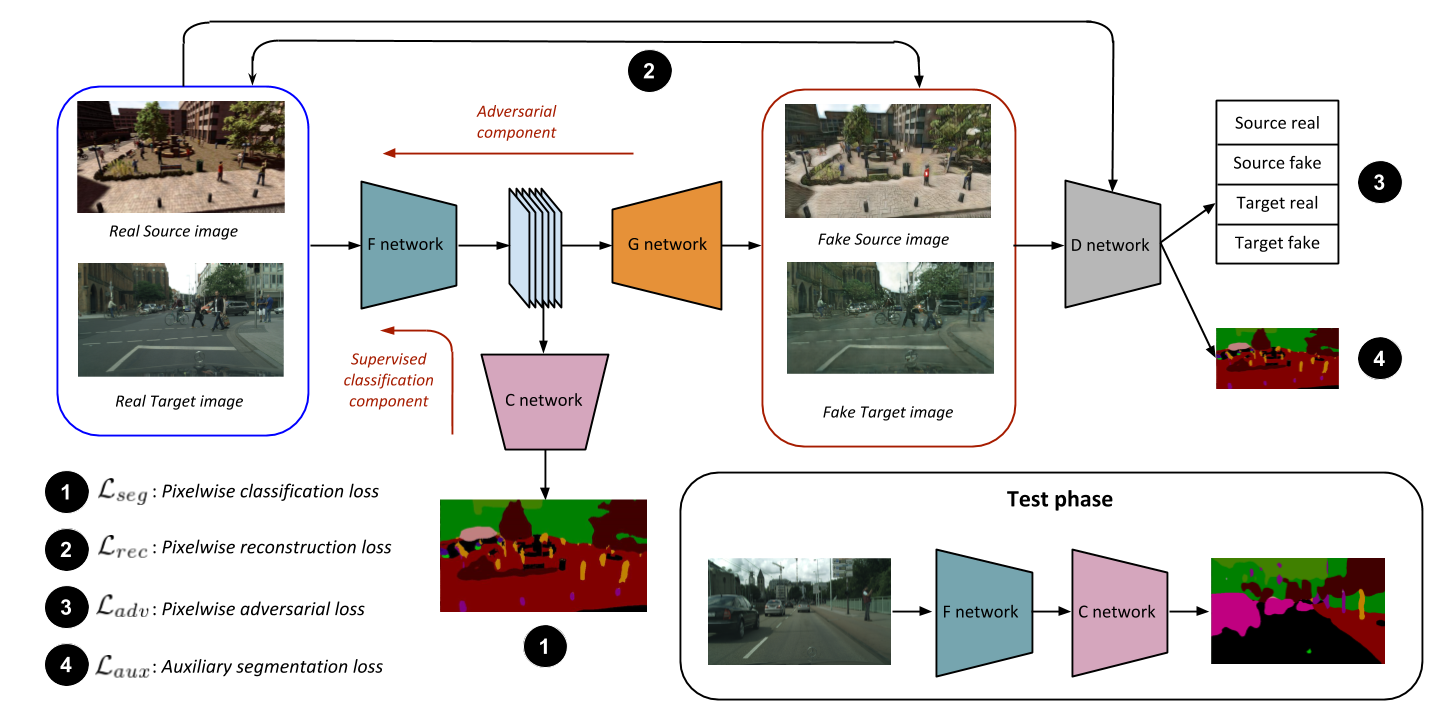}
\caption{During training, the $F$ and $C$ networks are trained jointly with the adversarial framework($G$-$D$ pair). $F$ is updated using a combination of supervised loss and an adversarial component. In the bottom right, we show the test time usage. Only the $F$ and $C$ network blocks are used. There is no additional overhead during evaluation compared to the base model.}
\label{fig:title}
\end{figure*}

Given a source image and label pair $\{X^s,Y^s\}$ as input, we begin by extracting a feature representation using the $F$ network. The classifier $C$ takes the embedding $F(X^s)$ as input and produces an image-sized label map $\hat{Y}^{s}$. The generator $G$ reconstructs the source input $X^{s}$ conditioned on the embedding. Following recent successful works on image generation, we do not explicitly concatenate the generator input with a random noise vector but instead use dropout layers throughout the $G$ network. As shown in Figure \ref{fig:title}, $D$ performs two tasks: (1) Distinguishing the real source input and generated source image as source-real/source-fake (2) producing a pixel-wise label map of the generated source image. 

Given a target input $X^t$, the generator network $G$ takes the target embedding from $F$ as input and reconstructs the target image. Similar to the previous case, $D$ is trained to distinguish between real target data (target-real) and the generated target images from $G$ (target-fake). However, different from the previous case, $D$ performs only a single task i.e. it classifies the target input as target-real/target-fake. Since the target data does not have any labels during training, the classifier network $C$ is not active when the system is presented with target inputs.

\subsection{Iterative optimization}
Fig.~\ref{fig:title} shows various losses used in our method. We begin by describing these losses, and then describe our iterative optimization approach.

The different adversarial losses used to train our models are shown in Table.~\ref{adv_loss}. In addition to these adversarial losses, we use the following losses: (1) $\mathcal{L}_{seg}$ and $\mathcal{L}_{aux}$ - pixel-wise cross entropy loss used in standard segmentation networks such as in FCN and (2) $\mathcal{L}_{rec}$ - $L_{1}$ loss between input and reconstructed images.

\begin{table*}[h]
\centering
\begin{tabular}{|c|c|c|}
\hline
 Type & Variants & Description \\ \hline
 & $\mathcal{L}^{s}_{adv,D}$ &  Classify real source input as \textit{src-real}; fake source input as \textit{src-fake} \\ \cline{2-3} 
 Within-domain & $\mathcal{L}^{s}_{adv,G}$ & Classify fake source input as \textit{src-real} \\ \cline{2-3} 
 & $\mathcal{L}^{t}_{adv,D}$ & Classify real target input as \textit{tgt-real}; fake target input as \textit{tgt-fake} \\ \cline{2-3} 
 & $\mathcal{L}^{t}_{adv,G}$ & Classify fake target input as \textit{tgt-real} \\ \hline
 Cross-domain & $\mathcal{L}^{s}_{adv,F}$ & Classify fake source input as real target (\textit{tgt-real}) \\ \cline{2-3}
 & $\mathcal{L}^{t}_{adv,F}$ & Classify fake target input as real source (\textit{src-real}) \\ \hline
\end{tabular}
\caption{Within-domain and Cross-domain adversarial losses that are used to update our networks during training. $G$ and $D$ networks are updated using only the within-domain losses while $F$ is updated only using the cross domain loss. All these adversarial losses originate from the $D$ network. $\mathcal{L}_{adv,X}$ implies that the gradients from the loss function $L$ are used to update $X$ only, while the other networks are held fixed.}
\label{adv_loss}
\end{table*}

The directions of flow of information across different network blocks are listed in Figure \ref{fig:explain}. In each iteration, a randomly sampled triplet $(X^s,Y^s,X^t)$ is provided to the system. Then, the network blocks are updated iteratively in the following order: 

\paragraph{(1) D-update:} For source inputs, $D$ is updated using a combination of within-domain adversarial loss $\mathcal{L}^{s}_{adv, D}$ and auxiliary classification loss $\mathcal{L}^{s}_{aux}$. For target inputs, it is updated using only the adversarial loss $\mathcal{L}^{t}_{adv, D}$. The overall loss $\mathcal{L}_{D}$ is given by $\mathcal{L}_{D} = \mathcal{L}^{s}_{adv, D} + \mathcal{L}^{t}_{adv, D} + \mathcal{L}^{s}_{aux}$.

\paragraph{(2) G-update:} In this step, the generator is updated using a combination of an adversarial loss $\mathcal{L}^{s}_{adv, G} + \mathcal{L}^{t}_{adv, G}$ intended to fool $D$ and a reconstruction loss $\mathcal{L}_{rec}$. The adversarial loss encourages realistic output from the generator. The pixelwise $L_1$ loss is crucial to ensure image fidelity between the generator outputs and the corresponding input images. The overall generator loss is given as: $\mathcal{L}_{G} = \mathcal{L}^{s}_{adv, G} + \mathcal{L}^{t}_{adv, G} +  \mathcal{L}^{s}_{rec} + \mathcal{L}^{t}_{rec}$.

\paragraph{(3) F-update:}
The update to the $F$ network is the critical aspect of our framework where the notion of domain shift is captured. The parameters of $F$ are updated using a combination of several loss terms: $\mathcal{L}_{F} = \mathcal{L}_{seg} + \alpha \, \mathcal{L}^s_{aux} + \beta \,(\mathcal{L}^s_{adv,F} + \mathcal{L}^{t}_{adv,F})$. As illustrated in Table \ref{adv_loss}, the adversarial loss terms used to update $F$ account for the domain adaptation. More specifically, the iterative updates described here can be considered as a min-max game between the $F$ and the $G$-$D$ networks. During the $D$ update step discussed earlier, the adversarial loss branch of $D$ learns to classify the input images as real or fake in a domain consistent manner. To update $F$, we use the gradients from $D$ that lead to a reversal in domain classification, i.e. for source embeddings, we use gradients from $D$ corresponding to classifying those embeddings as from target domain ($\mathcal{L}^s_{adv,F}$) and for target embeddings, we use gradients from $D$ corresponding to classifying those embeddings as from source domain ($\mathcal{L}^t_{adv,F}$). Note that, this is similar to the min-max game between the $G$-$D$ pair, except in this case, the competition is between classifying the generated image as from source/target domains instead of them being real/fake. 

\subsection{Motivating design choice of $D$}
\begin{itemize}
\item In traditional GANs that are derived from the DCGAN~\cite{dcgan2015} implementations, the output of the discriminator is a single scalar indicating the probability of the input being fake or drawn from an underlying data distribution. Recent works on image generation have utilized the idea of \textit{Patch} discriminator in which the output is a two dimensional feature map where each pixel carries a real/fake probability. This results in significant improvement in the visual quality of their generator reconstructions. We extend this idea to our setting by using a variant of the \textit{Patch} discriminator, where each pixel in the output map indicates real/fake probabilities across source and target domains hence resulting in four classes per pixel: \textit{src-real}, \textit{src-fake}, \textit{tgt-real}, \textit{tgt-fake}. 
\item In general, GANs are hard to train on tasks which involve realistic images of a larger scale. One promising approach to training stable generative models with the GAN framework is the Auxiliary Classifier GAN (AC-GAN) approach by Odena \emph{et al.} where they show that by conditioning $G$ during training and adding an auxiliary classification loss to $D$, they can realize a more stable GAN training and even generate large scale images. Inspired by their results on image classification, we extend their idea to the segmentation problem by employing an auxiliary pixel-wise labeling loss to the $D$ network. 
\end{itemize}
Both these components prove crucial to our performance. The ablation study performed in Section \ref{subsec:ablation} shows the effect of the above design choices on the final performance. Specific details about the architectures of these network blocks can be found in the supplementary material.

%% file: results.tex
\section{Experiments and Results}\label{sec:results}

In this section, we provide a quantitative evaluation of our method by performing experiments on benchmark datasets. We consider two challenging synthetic datasets available for semantic segmentation: SYNTHIA and GTA-5. SYNTHIA \cite{synthia} is a large dataset of photo-realistic frames rendered from a virtual city with precise pixel-level semantic annotations. Following previous works (~\cite{FCN_wild},~\cite{curriculum_DA}), we use the SYNTHIA-RAND-CITYSCAPES subset that contains $9400$ images with annotations that are compatible with cityscapes. GTA-5 is another large-scale dataset containing $24966$ labeled images. The dataset was curated by Richter \emph{et al.} \cite{GTA5} and is generated by extracting frames from the computer game Grand Theft Auto V. 

We used CITYSCAPES \cite{Cityscapes} as our real dataset. This dataset contains urban street images collected from a moving vehicle captured in $50$ cities around Germany and neighboring countries. The dataset comes with $5000$ annotated images split into three sets - $2975$ images in the \textit{train} set, $500$ images in the \textit{val} set and $1595$ images in the \textit{test} set. In all our experiments, for training our models we used labeled SYNTHIA or GTA-5 dataset as our source domain and unlabeled CITYSCAPES \textit{train} set as our target domain. We compared the proposed approach with the only two contemporary methods that address this problem: FCN in the wild~\cite{FCN_wild} and Curriculum Domain adaptation~\cite{curriculum_DA}. Following these approaches, we designate the 500 images from CITYSCAPES \textit{val} as our test set.

\paragraph{Architecture} In all our experiments, we used FCN-8s as our base network. The weights of this network were initialized with the weights of the VGG-16~\cite{VGGNet} model trained on Imagenet~\cite{imagenet}. 

\paragraph{Implementation details} In all our experiments, images were resized and cropped to $1024 \times 512$. We trained our model for $100,000$ iterations using Adam solver~\cite{adam_solver} with a batch size of $1$. Learning rate of $10^{-5}$ was used for $F$ and $C$ networks, and $2\times 10^{-4}$ for $G$ and $D$ networks. While evaluating on CITYSCAPES dataset whose images and ground truth annotations are of size $2048 \times 1024$, we first produce our predictions on the $1024 \times 512$ sized image and then upsample our predictions by a factor of $2$ to get the final label map, which is used for evaluation. Our training codes and additional results are publicly available. \footnote{Training code: \url{https://goo.gl/3Jsu2s}}


\subsection{SYNTHIA -> CITYSCAPES}
In this experiment, we use the SYNTHIA dataset as our source domain, and CITYSCAPES as our  target domain. We randomly pick $100$ images from the $9400$ labeled images of SYNTHIA dataset and use it for validation purposes, the rest of the images are used for training. We use the unlabeled images corresponding to the CITYSCAPES \textit{train} set for training our model. In order to ensure fairness of experimental results, we followed the exact evaluation protocol as specified by the previous works (~\cite{FCN_wild},\cite{curriculum_DA}):  The 16 common classes between SYNTHIA and CITYSCAPES are chosen used as our labels. The predictions corresponding to the other classes are treated as belonging to void class, and not backpropagated during training. The 16 classes are: sky, building, road, sidewalk, fence, vegetation, pole, car, traffic sign, person, bicycle, motorcycle, traffic light, bus, wall, and rider.

\begin{table*}
\begin{subtable}{1\textwidth}
\sisetup{table-format=-1.2}   
\centering
\resizebox{0.95\textwidth}{!}{
\begin{tabular}{c||c||c|c|c|c|c|c|c|c|c|c|c|c|c|c|c|c||c|c}
\hline 
\rule{0pt}{11ex}
Method & Base n/w &\rotatebox{90}{road} & \rotatebox{90}{sidewalk} & \rotatebox{90}{bldg} & \rotatebox{90}{wall} & \rotatebox{90}{fence} & \rotatebox{90}{pole} & \rotatebox{90}{t light} & \rotatebox{90}{t sign} & \rotatebox{90}{veg} & \rotatebox{90}{sky} & \rotatebox{90}{person} & \rotatebox{90}{rider} & \rotatebox{90}{car} & \rotatebox{90}{bus} & \rotatebox{90}{mbike} & \rotatebox{90}{bike} & \rotatebox{90}{mIOU} & \rotatebox{90}{mIOU gain} \\
\hline
\hline
\rule{0pt}{3ex}
Source only~\cite{FCN_wild} & Dilation-Frontend & 6.4 & 17.7 & 29.7 & 1.2 & 0.0 & 15.1 & 0.0 & 7.2 & 30.3 & 66.8 & 51.1 & 1.5 & 47.3 & 3.9 & 0.1 & 0.0 & 17.4 & \\
FCN wild~\cite{FCN_wild} & \cite{YuKoltun2016} & 11.5 & 19.6 & 30.8 & 4.4 & 0.0 & 20.3 & 0.1 & 11.7 & 42.3 & 
68.7 & 51.2 & 3.8 & 54.0 & 3.2 & 0.2 & 0.6 & 20.2 & 2.8\\

\hline
\rule{0pt}{3ex}

Source only~\cite{curriculum_DA} & FCN8s-VGG16 & 5.6 & 11.2 & 59.6 & 8.0 & 0.5 & 21.5 & 8.0 & 5.3 & 72.4 & 75.6 & 35.1 & 9.0 & 23.6 & 4.5 & 0.5 & 18.0 & 22.0 & \\
Curr. DA~\cite{curriculum_DA} & \cite{fcn2015} & 65.2 & 26.1 & 74.9 & 0.1 & 0.5 & 10.7 & 3.5 & 3.0 & 76.1 & 70.6 & 47.1 & 8.2 & 43.2 & 20.7 & 0.7 & 13.1 & 29.0 & 7.0 \\

\hline
\rule{0pt}{3ex}
Ours - Source only & FCN8s-VGG16 & 30.1 & 17.5 & 70.2 & 5.9 & 0.1 & 16.7 & 9.1 & 12.6 & 74.5 & 76.3 & 43.9 & 13.2 & 35.7 & 14.3 & 3.7 & 5.6 & 26.8 & \\
Ours - Adapted & \cite{fcn2015} & 80.1 & 29.1 & 77.5 & 2.8 & 0.4 & 26.8 & 11.1 & 18.0 & 78.1  & 76.7 & 48.2 & 15.2 & 70.5 & 17.4 & 8.7 & 16.7 & \textbf{36.1} & \textbf{9.3}\\
\hline
\hline
\rule{0pt}{3ex}
Target-only & FCN8s-VGG16 & 96.5 & 74.6 & 86.1 & 37.1 & 33.2 & 30.2 & 39.7 & 51.6 & 87.3  & 90.4 & 60.1 & 31.7 & 88.4 & 52.3 & 33.6 & 59.1 & 59.5 & -\\
\hline
\end{tabular}
}
\caption{SYNTHIA $\rightarrow$ CITYSCAPES}\label{tab:main_results_SYNTHIA}
\end{subtable}

\bigskip
\begin{subtable}{1\textwidth}
\sisetup{table-format=4.0} 
\centering
   \resizebox{\textwidth}{!}{
\begin{tabular}{c||c||c|c|c|c|c|c|c|c|c|c|c|c|c|c|c|c|c|c|c||c|c}
\hline 
\rule{0pt}{11ex}
Method & Base n/w & \rotatebox{90}{road} & \rotatebox{90}{sidewalk} & \rotatebox{90}{bldg} & \rotatebox{90}{wall} & \rotatebox{90}{fence} & \rotatebox{90}{pole} & \rotatebox{90}{t light} & \rotatebox{90}{t sign} & \rotatebox{90}{veg} & \rotatebox{90}{terrain} & \rotatebox{90}{sky} & \rotatebox{90}{person} & \rotatebox{90}{rider} & \rotatebox{90}{car} & \rotatebox{90}{truck} & \rotatebox{90}{bus} & \rotatebox{90}{train} & \rotatebox{90}{mbike} & \rotatebox{90}{bike} & \rotatebox{90}{mIOU} & \rotatebox{90}{mIOU gain} \\
\hline
\hline
\rule{0pt}{3ex}
Source only~\cite{FCN_wild} & Dilation-Frontend & 31.9 & 18.9 & 47.7 & 7.4 & 3.1 & 16.0 & 10.4 & 1.0 & 76.5 & 13.0 & 58.9 & 36.0 & 1.0 & 67.1 & 9.5 & 3.7 & 0.0 & 0.0 & 0.0 & 21.2 & \\
FCN wild~\cite{FCN_wild} & \cite{YuKoltun2016} & 70.4 & 32.4 & 62.1 & 14.9 & 5.4 & 10.9 & 14.2 & 2.7 & 79.2 & 
21.3 & 64.6 & 44.1 & 4.2 & 70.4 & 8.0 & 7.3 & 0.0 & 3.5 & 0.0 & 27.1 & 5.9\\

\hline
\rule{0pt}{3ex}

Source only~\cite{curriculum_DA} & FCN8s-VGG16 & 18.1 & 6.8 & 64.1 & 7.3 & 8.7 & 21.0 & 14.9 & 16.8 & 45.9 & 2.4 & 64.4 & 41.6 & 17.5 & 55.3 & 8.4 & 5.0 & 6.9 & 4.3 & 13.8 & 22.3\\
Curr. DA~\cite{curriculum_DA} & \cite{fcn2015} & 74.9 & 22.0 & 71.7 & 6.0 & 11.9 & 8.4 & 16.3 & 11.1 & 75.7 & 13.3 & 66.5 & 38.0 & 9.3 & 55.2 & 18.8 & 18.9 & 0.0 & 16.8 & 16.6 & 28.9 & 6.6\\

\hline
\rule{0pt}{3ex}
Ours - Source only & FCN8s-VGG16 & 73.5 & 21.3 & 72.3 & 18.9 & 14.3 & 12.5 & 15.1 & 5.3 & 77.2 & 17.4 & 64.3 & 43.7 & 12.8 & 75.4 & 24.8 & 7.8 & 0.0 & 4.9 & 1.8 & 29.6 & \\
Ours - Adapted & \cite{fcn2015} & 88.0 & 30.5 & 78.6 & 25.2 & 23.5 & 16.7 & 23.5 & 11.6 & 78.7  & 27.2 & 71.9 & 51.3 & 19.5 & 80.4 & 19.8 & 18.3 & 0.9 & 20.8 & 18.4 & \textbf{37.1} & 
\textbf{7.5}\\

\hline
\hline
\rule{0pt}{3ex}

Target-only & FCN8s-VGG16 & 96.5 & 74.6 & 86.1 & 37.1 & 33.2 & 30.2 & 39.7 & 51.6 & 87.3  & 52.6 & 90.4 & 60.1 & 31.7 & 88.4 & 54.9 & 52.3 & 34.7 & 33.6 & 59.1 & 57.6 & -\\

\hline
\end{tabular}
}
\caption{GTA5 $\rightarrow$ CITYSCAPES}\label{tab:main_results_GTA}
\end{subtable}
\caption{Results of Semantic Segmentation by adapting from (a) SYTNHIA to CITYSCAPES and (b) GTA-5 to CITYSCAPES. We compare with two approaches that use two different base networks. To obtain a fair idea about our performance gain, we compare with the Curriculum DA approach that uses the same base network as ours. The Target-only training procedure is the same for both the settings since in both cases the target domain is CITYSCAPES. However, the results in (a) are reported over the 16 common classes while the results in (b) are reported over all the 19 classes.}
\label{tab:all_results}
\end{table*}

Table~\ref{tab:main_results_SYNTHIA} reports the performance of our method in comparison with ~\cite{FCN_wild} and ~\cite{curriculum_DA}. The source-only model which corresponds to the no adaptation case i.e. training only using the source domain data achieves a mean IOU of 26.8.  The target-only values denote the performance obtained by a model trained using CITYSCAPES \textit{train} set (supervised training), and they serve as a crude upper bound to the domain adaptation performance. These values were included to put in perspective the performance gains obtained by the proposed approach. We observe that our method achieves a mean IOU of 36.1, thereby improving the baseline by \textbf{9.3 points}, thus resulting in a higher performance improvement compared to other reported methods.

\subsection{GTA5 -> CITYSCAPES}
In this experiment, we adapt from the GTA-5 dataset to the CITYSAPES dataset.  We randomly pick $1000$ images from the $24966$ labeled images of GTA-5 dataset and use it for validation purpose and use the rest of the images for training. We use the unlabeled images corresponding to the CITYSCAPES \textit{train} set for training our model. In order to ensure fairness of experimental results, we followed the exact evaluation protocol as specified by the previous works ( \cite{FCN_wild}, \cite{curriculum_DA}): we use 19 common classes between GTA-5 and CITYSCAPES as our labels. The results of this experiment are reported in Table.~\ref{tab:main_results_GTA}. Similar to the previous experiment, our baseline performance (29.6) is higher than the performance reported in  ~\cite{FCN_wild}, due to difference in network architecture and experimental settings. On top of this, the proposed approach yields an improvement of 7.5 points to obtain a mIOU of \textbf{37.1}. This performance gain is higher than that achieved by the other compared approaches.

\textbf{Note regarding different baselines:} The baseline numbers reported by us do not match with the ones reported in ~\cite{curriculum_DA} and ~\cite{FCN_wild} due to different experimental settings (this mismatch was also reported in ~\cite{curriculum_DA}). However, we would like to point out that we improve over a stronger baseline compared to the other two methods in both our adaptation experiments. In addition, ~\cite{curriculum_DA} uses additional data from PASCAL-CONTEXT \cite{pascal-context} dataset to obtain the superpixel segmentation. In contrast, our approach is a single stage end-to-end learning framework that does not use any additional data and yet obtains better performance improvement.

%% file: ablation_studies.tex
\section{Discussion}\label{sec:discuss}
In this section, we perform several exploratory studies to give more insight into the functionality and effectiveness of the proposed approach. similar to the previous section, all the evaluation results are reported on the CITYSCAPES \textit{val} set, unless specified otherwise. We denote this set as the \textit{test set}. 


\subsection{Effect of Image Size}
The datasets considered in this paper consists of images of large resolution which is atleast twice larger than the most commonly used Segmentation benchmarks for CNNs i.e. PASCAL VOC (500$\times$300) and MSCOCO (640$\times$480). In this setting, it is instructive to understand the effect of image size on the performance of our algorithm both from a quantitative and computational perspective. Table \ref{tab:img_size} presents the results of our approach applied over three different image sizes along with the training and evaluation times. It should be noted that the Curriculum DA approach \cite{curriculum_DA} used a resolution of 640$\times$320. By comparing with our main results in Table \ref{tab:main_results_SYNTHIA}, we see that our approach provides a higher relative performance improvement over a similar baseline.

\begin{table}[!h]
\caption{Mean IoU values and computation times across different image size on the SYNTHIA $\rightarrow$ CITYSCAPES setting. The numbers in \textbf{bold} indicate the absolute improvement in performance over the \textit{Source-only} baseline. The reported training and evaluation times are for the proposed approach and are averaged over training and evaluation runs.}
\label{tab:img_size}
\centering
\resizebox{0.45\textwidth}{!}{
\begin{tabular}{l|c|c|c}
\hline
Image size & $512 \times 256$ & $640 \times 320$ & $1024 \times 512$ \\   
\hline
 mIOU-\textit{Source-only} & 21.5 & 23.2 & 26.8 \\ 
 mIOU-\textit{Ours} & 31.3 (\textbf{+9.8}) & 34.5 (\textbf{+11.3}) & 36.1 (\textbf{+9.3}) \\ 
 Train time (per image) & 1.5s & 2.1s & 2.9s \\
 Eval time (per image) & 0.16s & 0.19s & 0.3s \\
 \end{tabular}
 }
\end{table}

\subsection{Comparison with direct style transfer}

Generative methods for style transfer have achieved a great amount of success in the recent past. A simple approach to performing domain adaptation is to use such approaches as a data augmentation method: transfer the images from the source domain to target domain and use the provided source ground truth to train a classifier on the combined source and target data. In order to compare the proposed approach with this direct data augmentation procedure, we used a state of the art generative approach (CycleGAN \cite{cycleGAN_ICCV}) to transfer images from source domain to target domain. From our experiment, using generative approaches solely as a data augmentation method provides only a relatively small improvement over the source-only baseline and clearly suboptimal compared to the proposed approach. However, as shown in a recent approach by Hoffman \emph{et al.} \cite{cycada}, such cross domain transfer can be performed by a careful training procedure. The results obtained by the proposed approach is comparable or better then \cite{cycada}. Combining both approaches to produce a much stronger domain adaptation technique for segmentation is under progress.



\subsection{Component-wise ablation}\label{subsec:ablation}

In this experiment, we show how each component in our loss function affects the final performance. We consider the following cases: (a) Ours(full): the full implementation of our approach (b) Ours w/o auxiliary pixel-wise loss: Here, the output of the $D$ network is a single branch classifying the input as real/fake. This corresponds to $\alpha=0$ in the $F$-update step. Note that, setting both $\alpha$ and $\beta$ as zero corresponds to the source-only setting in our experiments. Setting only $\beta=0$ does not improve over the source-only baseline as there is no cross domain adversarial loss. (c) Ours w/o \textit{Patch} discriminator: Instead of using the $D$ network as a \textit{Patch} discriminator, we used a regular GAN-like discriminator where the output is a 4-D probability vector that the input image belongs to one of the four classes - \textit{src-real}, \textit{src-fake}, \textit{tgt-real} and \textit{tgt-fake}. (d) Feature space based $D$: In this setting, we remove the $G$-$D$ networks and apply an adversarial loss directly on the embedding. This is similar to the global alignment setting in the FCN-in-the-wild approach \cite{FCN_wild}.  

The mean IoU results on the \textit{test set} are shown in Table. \ref{tab:ablation}. It can be observed that each component is very important to obtain the full improvement in performance. 

\begin{table}[!h]
\caption{Ablation study showing the effect of each component on the final performance of our approach on the SYNTHIA $\rightarrow$ CITYSCAPES setting}
\label{tab:ablation}
\centering
\begin{tabular}{l|c}
\hline
Method & \text{mean IoU} \\   
\hline
 Source-only & 22.2 \\ 
Feature space based $D$ & 25.3\\ 
 Ours w/o \textit{Patch} Discriminator & 28.3 \\
 Ours w/o auxiliary loss ($\alpha=0$) & 29.2 \\
 Ours & \textbf{34.5}\\
 \end{tabular}
\end{table}

\subsection{Cross Domain Retrieval}\label{subsec:retrieval}
A crucial aspect of domain adaptation is in finding good measures of domain discrepancy that provide a good illustration of the domain shift. While there exist several classical measures such as $\mathcal{A}$-distance \cite{ben2007bounds} and $MMD$ \cite{mmd2011} for the case of image classification, the extension of such measures for a pixel-wise problem such as semantic segmentation is non-trivial. In this section, we devise a simple experiment in order to illustrate how the proposed approach brings source and target distributions closer in the learnt embedding space. We start with the last layer of the $F$ network, which we label as the embedding layer, whose output is a spatial feature map. We perform an average pooling to reduce this spatial map to a 4096 dimensional feature descriptor for each input image. 

We begin the cross domain retrieval task by choosing a pool of $N = N_{src} + N_{tgt}$ images from the combined source and target training set. Let $X$ denote these set of images and $F_{X}$ denote the set of the feature descriptors computed for $X$. Then, we choose two query sets, one consisting of source images ($S$) and the other consisting of target images ($T$), each disjoint with $X$. Let the corresponding feature sets be denoted as $Q_{S}$ and $Q_{T}$. We retrieve k-NN lists for each item in the query set from the combined feature set $F_{X}$. For each query point in $Q_{S}$, we count the number of target samples retrieved in the corresponding k-NN list. $|A_k|$ indicates the average number of target samples retrieved over the entire source query set $Q_{S}$. For each query point in $Q_{T}$, we count the number of source samples retrieved in the corresponding k-NN list. $|B_k|$ indicates the average number of source samples retrieved over the entire target query set $Q_{T}$. We used cosine similarity as a metric to compute the k-NN lists. If more target samples are retrieved for a source query point (and vice-versa), it suggests that source and target distributions are aligned well in the feature space.

\begin{figure}
\centering
\begin{subfigure}{.25\textwidth}
  \centering
  \includegraphics[width=\linewidth]{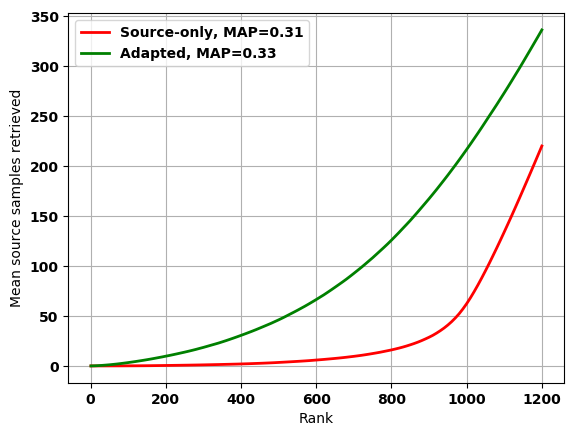}
  \caption{Target $\rightarrow$ Source, $|B_{k}|$ (vs) $k$}
  \label{fig:a}
\end{subfigure}%
\begin{subfigure}{.25\textwidth}
  \centering
  \includegraphics[width=\linewidth]{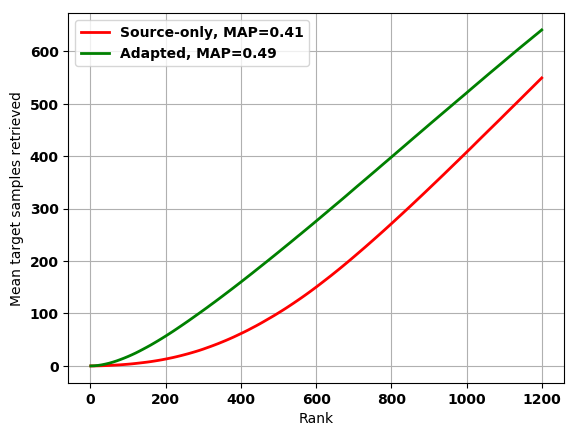}
  \caption{Source $\rightarrow$ Target, $|A_{k}|$ (vs) $k$}
  \label{fig:b}
\end{subfigure}
\caption{Illustration of Domain Adaptation achieved by the proposed approach. The plot compares the average number of retrieved sampled for the cross domain retrieval task described in Section \ref{subsec:retrieval} between the \text{source-only} model and the model adapted using the proposed approach. Target $\rightarrow$ Source implies that the query set used belongs to target domain ($Q_{T}$) and items queried for from the set $X$ belong to the source domain and vice-versa for Source $\rightarrow$ Target. In general, the values plotted on the y-axis corresponds to the number of samples retrieved from the set $X$ that belong to the opposite domain as to that of the query set.}
\label{fig:retrieval}
\end{figure}

For this experiment, the sizes of query sets and the feature set $F_{X}$ are as follows: $N_{src}$ = $N_{tgt}$ = 1000, $|Q_{S}|$ = 1000, $|Q_{T}|$ = 1000. The mean average precision (mAP) was computed across the entire query sets for the respective cross domain tasks. Figure \ref{fig:retrieval} shows the plot of the quantities $|A_k|$ (Fig.\ref{fig:b}) and $|B_k|$ (Fig.\ref{fig:a}) for a range of values of $k$. It can be observed from the plots in both the tasks that for any given rank $k$, the number of cross domain samples retrieved by the adapted model is higher than the source-only model. This effect becomes more clear as $k$ increases. This observation is supported by better mAP values for the adapted model as shown in Figure \ref{fig:retrieval}. While this by itself is not a sufficient condition for better segmentation performance, however this along with the results from Table \ref{tab:all_results} imply that the proposed approach performs domain adaptation in a meaningful manner. Owing to the difficulty in visualizing the mapping learned for segmentation tasks, a cross domain retrieval experiment can be seen as a reasonable measure of how domain gap is reduced in the feature space.




\subsection{Generalization to unseen domains}
A desirable characteristic of any domain adaptation algorithm is domain generalization i.e. improving performance over domains that are not seen during training. To test the generalization capability of the proposed approach, we test the model trained for the SYNTHIA $\rightarrow$ CITYSCAPES setting on the CamVid dataset \cite{camvid2009}. We choose to evaluate our models on the 10 common classes among the three datasets. Table \ref{tab:camvid} shows the mean IoU values computed for the source-only baseline and the adapted model. The proposed approach yields a raw improvement of \textbf{8.3} points in performance which is a significant improvement considering the fact that CamVid images are not seen by the adapted model during training. This experiment showcases the ability of the proposed approach to learn domain invariant representations in a generalized manner.

\begin{table}[h!]
\caption{Mean IoU segmentation performance measured on a third unseen domain (CamVid dataset) for the models corresponding to the SYNTHIA $\rightarrow$ CITYSCAPES setting}
\label{tab:camvid}
\centering
\begin{tabular}{l|c}
\hline
Method & \text{mean IoU} \\   
\hline
 Source-only & 36.1 \\  
 Ours & \textbf{44.4}\\
 \end{tabular}
\end{table}
\vspace{-2mm}

%% file: conclusion.tex
\section{Conclusion and Future Work}\label{sec:conclusion}
In this paper, we have addressed the problem of performing semantic segmentation across different domains. In particular, we have considered a very hard case where abundant supervisory information is available for synthetic data (source) but no such information is available for real data (target). We proposed a joint adversarial approach that transfers the information of the target distribution to the learned embedding using a generator-discriminator pair. We have shown the superiority of our approach over existing methods that address this problem using experiments on two large scale datasets thus demonstrating the generality and scalability of our training procedure. Furthermore, our approach has no extra computational overhead during evaluation, which is a critical aspect when deploying such methods in practice. As future work, we would like to extend this approach to explicitly incorporate geometric constraints accounting for perspective variations and to adapt over temporal inputs such as videos across different domains.

%% file: ackno.tex
\section{Acknowledgement}
The Authors acknowledge support of the following organisations for sponsoring this work: (1) Avitas Systems, a GE venture (2) MURI from the Army Research Office under the Grant No. W911NF-17-1-0304. This is part of the collaboration between US DOD, UK MOD and UK Engineering and Physical Research Council (EPSRC) under the Multidisciplinary University Research Initiative.